\documentclass[10pt,twocolumn,letterpaper]{article}

\usepackage[pagenumbers]{iccv} % To force page numbers, e.g. for an arXiv version

\usepackage{adjustbox}
\usepackage{amsfonts}
\usepackage{pifont}
\newcommand{\cmark}{\ding{51}}%
\newcommand{\xmark}{\ding{55}}%

\definecolor{iccvblue}{rgb}{0.21,0.49,0.74}

\usepackage[pagebackref,breaklinks,colorlinks,allcolors=iccvblue]{hyperref}
\usepackage{amssymb}
\usepackage{amsmath}

\usepackage{multirow}
\usepackage{adjustbox}
\usepackage{booktabs}
\usepackage{multirow}
\usepackage{array}
\usepackage{makecell}
\usepackage{xcolor}

\def\paperID{13369} % *** Enter the Paper ID here
\def\confName{ICCV}
\def\confYear{2025}

\title{Vector Quantized Feature Fields for Fast 3D Semantic Lifting}

\author{George Tang\\
MIT\\
\and
Aditya Agarwal\\
MIT\\
\and
Weiqiao Han\\
MIT\\
\and
Trevor Darrell\\
UC Berkeley
\and 
Yutong Bai\\
UC Berkeley
}

\begin{document}
\twocolumn[{%
\maketitle
\begin{center}
    \captionsetup{type=figure}
    \includegraphics[width=\textwidth]{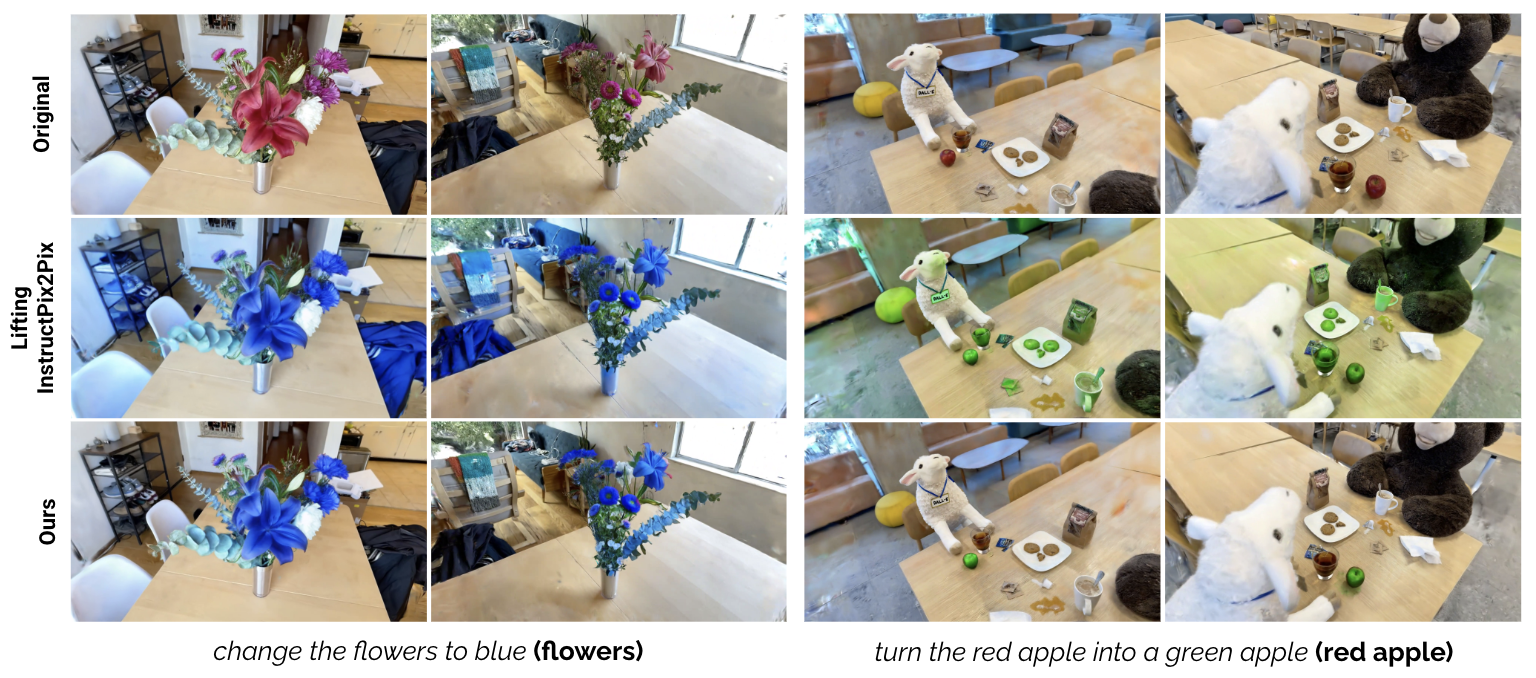}
    \captionof{figure}{We compare lifting with semantic lifting of per-view InstructPix2Pix edits for object-centric 3D editing. Naive lifting often corrupts other parts of the scene. Semantic lifting eliminates this issue without added overhead by integrating per-view localization masks instantly determined via a Vector Quantized Feature Field of the scene. \textit{Italicized} represents the InstructPix2Pix prompt, while \textbf{(bold)} denotes the localization mask prompt. The displayed images are novel views of the edited scene lifted via Gaussian Splatting.}
    \label{splash}
\end{center}
}]

\begin{abstract}
We generalize lifting to semantic lifting by incorporating per-view masks that indicate relevant pixels for lifting tasks. These masks are determined by querying corresponding multiscale pixel-aligned feature maps, which are derived from scene representations such as distilled feature fields and feature point clouds. However, storing per-view feature maps rendered from distilled feature fields is impractical, and feature point clouds are expensive to store and query. To enable lightweight on-demand retrieval of pixel-aligned relevance masks, we introduce the Vector-Quantized Feature Field. We demonstrate the effectiveness of the Vector-Quantized Feature Field on complex indoor and outdoor scenes. Semantic lifting, when paired with a Vector-Quantized Feature Field, can unlock a myriad of applications in scene representation and embodied intelligence. Specifically, we showcase how our method enables text-driven localized scene editing and significantly improves the efficiency of embodied question answering.
\end{abstract}

\begin{table*}
    \centering
    \adjustbox{max width=\linewidth}{
    \begin{tabular}{lcccc}
    \toprule
        Method & 3D Consistent Features & Instant Query & Low Memory Usage & Suitable for Lifting Tasks \\
    \midrule
        2D Visual Grounding & \xmark & \xmark & \cmark & depends on consistency \\
        Feature Fields & \cmark & \xmark & \cmark & \cmark \\
        Feature Splats & \cmark & \xmark & \xmark & \cmark \\
        Feature Point Clouds &  \cmark &  \cmark & \xmark & \xmark \\
        Ours &  \cmark &  \cmark &  \cmark &  \cmark \\
    \bottomrule
    \end{tabular}
    } % end adjustbox
    \caption{\textbf{Comparison of 3D scene representation methods for semantic lifting.} Our Vector-Quantized Feature Field (VQ-FF) is optimal across all criteria required by semantic lifting: features are 3D consistent, instant query capability, memory efficiency, and suitability for lifting tasks.}

    \label{tab:method_criteria}
\end{table*}

\section{Introduction}
Recent breakthroughs in 2D vision language foundation models \cite{radford2021learningtransferablevisualmodels, caron2021emergingpropertiesselfsupervisedvision, oquab2024dinov2learningrobustvisual,kirillov2023segment, ravi2024sam2segmentimages} have greatly improved zero-shot capability. These advances raise an exciting prospect for 3D scenes: lifting the outputs of these models applied to image sequences into 3D for applications ranging from text-driven scene editing to scene understanding. While lifting RGB for novel view rendering as well as 2D features such as Clip and Dino into a queryable 3D representation has been successful \cite{shen2023distilledfeaturefieldsenable, kerr2023lerflanguageembeddedradiance, qin2024langsplat3dlanguagegaussian, wu2024opengaussianpointlevel3dgaussianbased, zhou2024feature3dgssupercharging3d, t2024lift3dzeroshotlifting2d}, scaling this process to more sophisticated tasks, such as object-centric editing or embodied question answering (EQA), reveals significant bottlenecks in task performance and scalability. For example, InstructNeRF2NeRF \cite{haque2023instructnerf2nerfediting3dscenes} builds on InstructPix2Pix \cite{brooks2023instructpix2pixlearningfollowimage} to enable text-driven scene edits, but the underlying 2D editing model often introduces extraneous modifications that spill over to areas not mentioned in the prompt. Similarly, current EQA methods \cite{majumdar2023openeqa} frequently process the entire sequence of images in large batches due to not being able to efficiently determine which frames are relevant.

To remedy this, we introduce the notion of semantic lifting, which generalizes lifting to incorporate per-image masks indicating which parts of each image are relevant. Semantic lifting allows us to reformulate traditional lifting tasks to be more efficient. The user can determine the masks with prompting e.g. text or feature embedding query.

However, computing these per-image masks is not straightforward. A naive approach might construct a dense point cloud of 3D-consistent features \cite{jatavallabhula2023conceptfusionopensetmultimodal3d, peng2023openscene3dsceneunderstanding} from view-inconsistent 2D feature maps, or learn a 3D feature field \cite{shen2023distilledfeaturefieldsenable, kerr2023lerflanguageembeddedradiance, qin2024langsplat3dlanguagegaussian} from 2D feature maps, then use the feature field to render pixel-aligned view-consistent feature maps for each image to identify relevant regions. Unfortunately, dense point clouds with thousands to millions of points, each storing high-dimensional feature vectors, can consume dozens of GBs. Moreover, even with tens of millions of points, point clouds are still not pixel-aligned and cannot be directly integrated into the lifting process. 3D feature fields like LERF \cite{kerr2023lerflanguageembeddedradiance} and LangSplat \cite{qin2024langsplat3dlanguagegaussian} fare better on memory in principle. However, they still require iterating over numerous 3D feature scales and performing view-dependent renders to obtain a relevance mask for each camera frame, which becomes expensive in large scenes or repeated user interactions.

Our approach sidesteps these obstacles by, interestingly, lifting. First, we lift a compact 3D-consistent feature representation via a 3D feature field. Then, we lift the corresponding rendered feature maps into our proposed Vector-Quantized Feature Field (VQ-FF) that requires drastically less memory while preserving the semantic fidelity necessary for identifying relevant pixels in each view. We showcase how, using the VQ-FF, we can obtain the masks for semantic lifting with zero overhead compared to the original lifting formulation for the task.

Our key insight is that we can quantize the feature maps over all images into a compact codebook, which we can query to determine image masks instantly. We achieve this via a local-global decomposition strategy: first, we apply local quantization on each feature map using superpixel partitions, which helps align quantization boundaries with the underlying image structure and also denoises spurious features; then, we perform global quantization across the entire scene to form a unified codebook. This two-stage procedure drastically compresses the feature maps while preserving enough semantic information to retrieve pixel-aligned relevance masks on demand. Furthermore, it makes the VQ-FF construction feasible, as we explain later.

To summarize, our contributions are: 
\begin{itemize}
    \item We generalize 3D lifting to include per-image masks, enabling instant object detection, targeted edits, and efficient EQA.
    \item We introduce a novel pipeline for compressing multiscale feature maps into a single, compact VQ-FF representation that yields instant queries via local/global quantization that suits the problem formulation constraints.
    \item We evaluate our system on a diverse set of scenes from the LERF dataset, demonstrating that our compact representation preserves or improves query fidelity while enabling semantic lifting at scale.
\end{itemize}
\begin{figure*}[t]
    \centering
    \includegraphics[width=\textwidth]{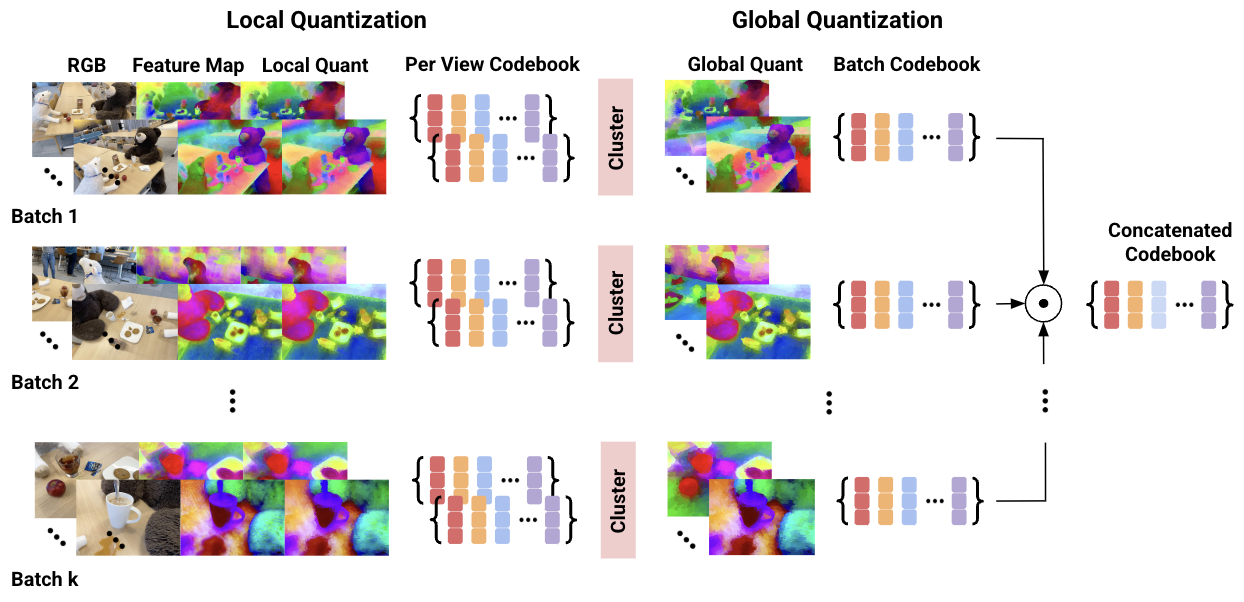}
    \caption{We first split the image sequence into $k$ batches. Within each batch, we perform local quantization by rendering feature maps and quantizing based on superpixels, followed by global quantization that employs clustering to quantize within a batch. Batching takes advantage of the structure and feature similarity between consecutive images in the sequence and greatly reduces the runtime of global quantization. We do not quantize over scales due to how relevancy computation is formulated.}
    \label{fig:method.png}
\end{figure*}

\section{Related Work}
\label{sec:related_work}
\textbf{Queryable Scene Representations}
2D foundation models such as Clip \cite{radford2021learningtransferablevisualmodels}, LSeg \cite{li2022languagedrivensemanticsegmentation} and Dino \cite{caron2021emergingpropertiesselfsupervisedvision, oquab2024dinov2learningrobustvisual}, produce semantically rich and pixel-aligned embeddings, making them useful for various downstream queries. For image sequences, per-image 2D features are not 3D consistent and relying on them for 3D tasks may introduce artifacts. Extending these embeddings to 3D, works like ConceptFusion \cite{jatavallabhula2023conceptfusionopensetmultimodal3d} and OpenScene \cite{peng2023openscene3dsceneunderstanding} combined foundation model-derived features with point clouds to construct 3D consistent feature representations for scene querying. However, point cloud-based methods are memory intensive. ConceptFusion, for instance, constructs dense point clouds with around $10^7$ points and a 1024-dimensional embedding per point, resulting in approximately 40GB of memory usage per scene. Furthermore, the points are not pixel-aligned, hampering direct integration with lifting formulations.

Distilling feature fields into learned scene representations is an alternative for constructing 3D consistent representations directly from posed RGB images \cite{shen2023distilledfeaturefieldsenable, kerr2023lerflanguageembeddedradiance, qin2024langsplat3dlanguagegaussian, wu2024opengaussianpointlevel3dgaussianbased, zhou2024feature3dgssupercharging3d, t2024lift3dzeroshotlifting2d, tang2024segmentmeshzeroshotmesh}. For example, LERF distills multiscale Clip features into a NeRF to obtain 3D consistent features, which requires less memory than ConceptFusion but is expensive to perform scene-level queries. For example, determining if an object exists in a scene requires rendering over many feature maps. Furthermore, LERF features are often noisy \cite{qin2024langsplat3dlanguagegaussian}. Gaussian Splatting plus Segment Anything (SAM) \cite{kirillov2023segment, ravi2024sam2segmentimages} -based methods such as LangSplat and OpenGaussian \cite{wu2024opengaussianpointlevel3dgaussianbased} reduce the number of scales to make feature fields feasible for Gaussian Splatting and reduce noise. However, they cannot handle granularity below SAM.

\noindent\textbf{3D Lifting}
A parallel thread of research focuses on lifting information from 2D into 3D. NeRF \cite{mildenhall2020nerfrepresentingscenesneural}, Gaussian Splatting \cite{kerbl20233dgaussiansplattingrealtime}, and Distilled Feature Fields \cite{shen2023distilledfeaturefieldsenable} can be formulated as lifting RGB into a consistent 3D scene representation. Works such as Panoptic Lifting \cite{siddiqui2022panopticlifting3dscene}, Contrastive Lifting \cite{bhalgat2023contrastivelift3dobject}, and 3DIML \cite{tang2024efficient3dinstancemapping} aim to fuse 2D inconsistent semantic and instance segmentation into a consistent 3D segmentation label field. Techniques like TripoSR \cite{tochilkin2024triposrfast3dobject}, InstantMesh \cite{xu2024instantmeshefficient3dmesh}, SceneComplete \cite{agarwal2024scenecompleteopenworld3dscene}, and Trellis \cite{xiang2024structured3dlatentsscalable} tackle 3D reconstruction via recovering high-quality geometry creation from a single or multiple views. Meanwhile, methods such as InstructNeRF2NeRF enable instruction-based editing for 3D scenes, e.g. ``remove the chair" or ``make the table smaller". Methods used for EQA explore different ways of querying foundation models (specifically LLMs and VLMs) to produce a response for 3D understanding without additional fine-tuning~\cite{majumdar2023openeqa}.

% feed an image sequence along with a question into a vision-language model (VLM) to produce a response for 3D understanding~\cite{majumdar2023openeqa}.

Naive lifting often leads to issues. For example, as in Figure \ref{splash}, naive lifting of InstructPix2Pix \cite{brooks2023instructpix2pixlearningfollowimage}, often results in extraneous edits not relevant to the semantic region referenced in the prompt. OpenEQA \cite{majumdar2023openeqa} inputs a subset of the scan sequence into a VLM for EQA, wasting tokens on frames unrelated to the query. In the next section, we describe how we address the above challenges by generalizing to semantic lifting with zero overhead via VQ-FF.

\section{Semantic Lifting}
Suppose we are given a sequence of images and corresponding poses $\{x_i, p_i\}$. We define lifting as
\begin{align}
R = L(\{f(x_i), p_i\})
\end{align}
where $L$ is the lifting function, $f$ is a per-image transform function, and $R$ is the output. To address the shortcomings of lifting, we introduce a more general form, semantic lifting. Semantic lifting incorporates a per-image binary mask $m_i$ that indicates which part of the image is relevant, which can be useful for localized editing and grounding tasks.
\begin{align}
R = L(\{m_i \odot f(x_i), p_i\})
\end{align}
where $\odot$ denotes masking operations \footnote{The simplest masking operation is bit masking}. Semantic lifting generalizes lifting since we can set $m_i$ to the identity mask. In our work, we compute masks via
\begin{align}
m_i = h(g(p_i), \text{prompt}) > \tau
\end{align}
where $g$, called the grounding scene representation, produces view-consistent feature maps for the given poses, and $h(\cdot, \text{prompt})$ denotes querying the feature map with the prompt. This produces per-view \emph{relevancy maps} that are thresholded to get the masks. In practical applications, $g$ is also constructed via lifting view-inconsistent feature maps produced by 2D foundation models such as Clip or Dino into 3D. It takes the form of a learned scene representation or point cloud, as described in Section~\ref{sec:related_work}, and thus inherits the pros and cons of its underlying structure.

\begin{figure}[t]
    \centering
    \includegraphics[width=0.475\textwidth]{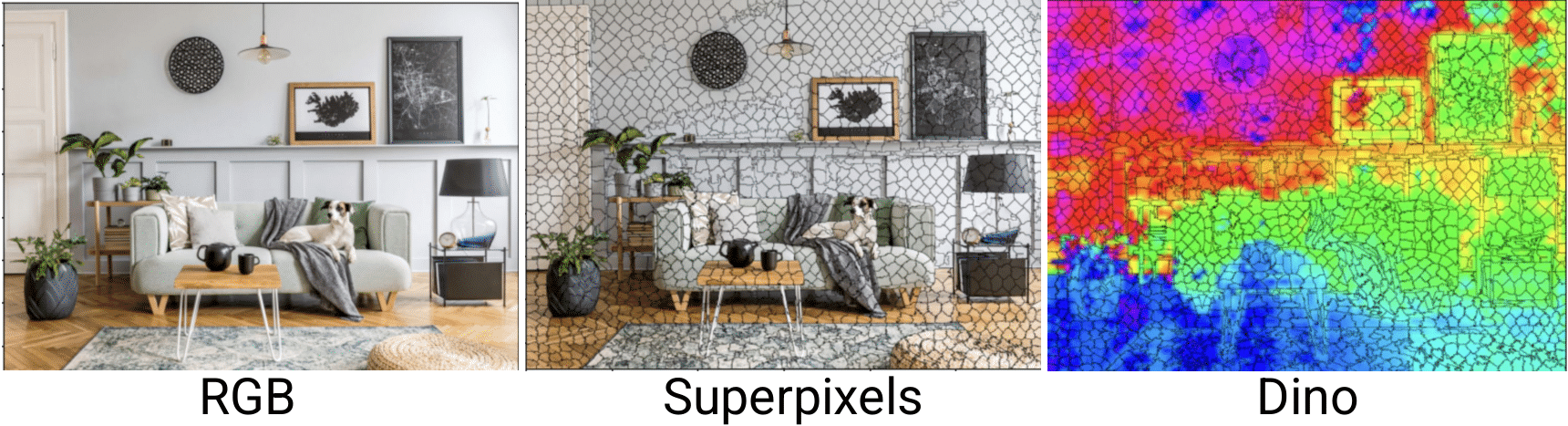}
    \caption{Visual foundation features such as Dino are well aligned with superpixels boundaries.}
    \label{fig:feature_superpixel_alignment}
\end{figure}

We choose distilled feature fields, e.g. LERF or LangSplat, as the grounding scene representation due to their compactness and ease of construction. Then $g(\cdot) = \text{render}(\cdot)$ and $h(\cdot)$ corresponds to the following
\begin{align}\label{eq:relevancy_map}
h_i = \min_i \frac{\exp(\phi_{\text{lang}} \cdot \phi_{\text{quer}})}{\exp(\phi_{\text{lang}} \cdot \phi_{\text{canon}}^i) + \exp(\phi_{\text{lang}} \cdot \phi_{\text{quer}})}
\end{align}
where $\phi_{\text{lang}} = \text{encode}(\text{prompt})$ and $\phi_{\text{canon}}^i = \text{encode}(s_i)$ for some encode function, e.g. Clip text encoder, and $s_i$ is a canonical phrase $\in$ (``object", ``things", ``stuff", and ``texture"). For a discussion of the applications of semantic lifting compared to lifting, see Section 4.3.

\section{Vector Quantized Feature Fields}
\subsection{Motivation}
Naively querying each view of a distilled feature field given a prompt is expensive, and it is infeasible to store all feature maps of the image sequence in memory. We showcase how we can obtain the masks needed for semantic lifting with \emph{zero overhead} besides construction.

Assuming infinite memory, we can store the feature maps and run all our queries in $O(1)$. However, for a length $N$ sequence for multiscale $M \times H \times W$ feature maps of dimension $D$, the total memory consumed is $O(NMHWD)$, which for LERF can reach up to 16TB of data and Langsplat 1.6TB. In fact, a single feature map for LERF is about 40GB. Thus, to compute \Cref{eq:relevancy_map} per image, each scale is rendered in succession. Rendering takes $O(NM\text{render}(H, W, D))$ time, where $\text{render}$ denotes the time complexity of the rendering function.

For LERF, $\text{render}(H, W, D) = \frac{HWD}{B}$ where $B$ is dependent on GPU memory, i.e. LERF batches pixel-level volumetric ray accumulations. LangSplat is more complex and depends on the Gaussian Splatting setup.

To reduce the query time and memory usage, we construct a Vector Quantized Feature Field, an approximation of the feature maps rendered from the 3D feature field. Instead of storing feature maps, it constructs feature codebook $C$ and stores corresponding codebook index maps ${V}_i$ in place of the feature maps \footnote{We quantize each scale separately due to \Cref{eq:relevancy_map}}. The index maps use $O(NMHW)$ memory and codebook $O(KD)$, where $K$ is the size of the codebook. We aim for $K = \frac{NMHW}{D}$ to balance, resulting in $O(NMHW)$ memory usage. We use \Cref{eq:relevancy_map} to determine a relevancy mask. Given a text query, finding relevant embeddings based on \Cref{eq:relevancy_map} and then finding the corresponding relevant pixels in ${V}_i$ is $O(KD)$ and $O(NHW)$, and can be trivially vectorized or parallelized, taking $O(1)$ time. Table \ref{tab:method_criteria} shows that VQ-FF is the only method that achieves all desirable properties that can support fast semantic lifting.

\begin{figure*}[t]
    \centering
    \includegraphics[width=0.95\textwidth]{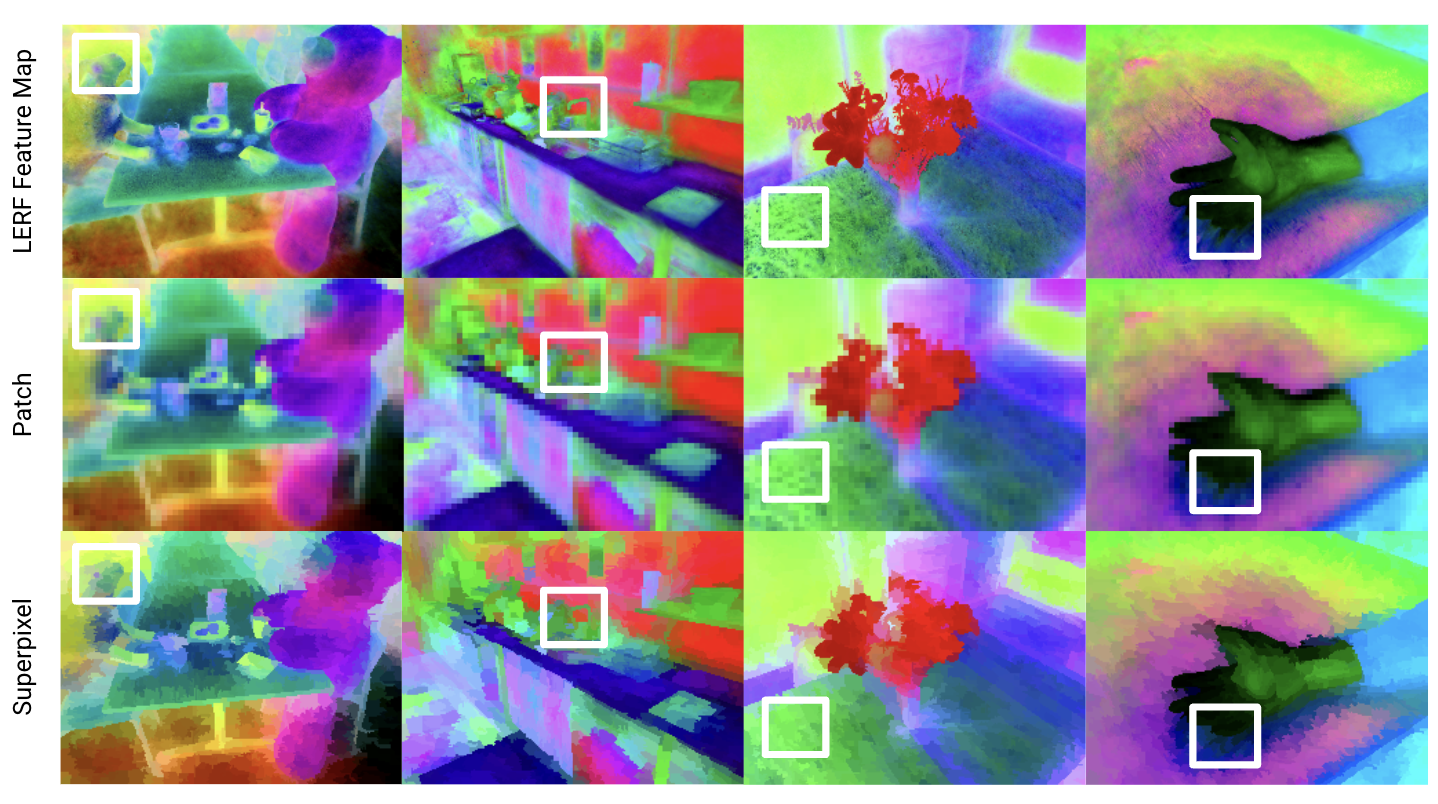}
    \caption{Visual comparison of superpixel-based vs patch-based local quantization feature map reconstruction quality. The white box highlights our method's denoising capability while preserving important structural details.}
    \label{fig:feature_quant_local_lerf}
\end{figure*}

\begin{figure*}[t]
    \centering
    \includegraphics[width=0.95\textwidth]{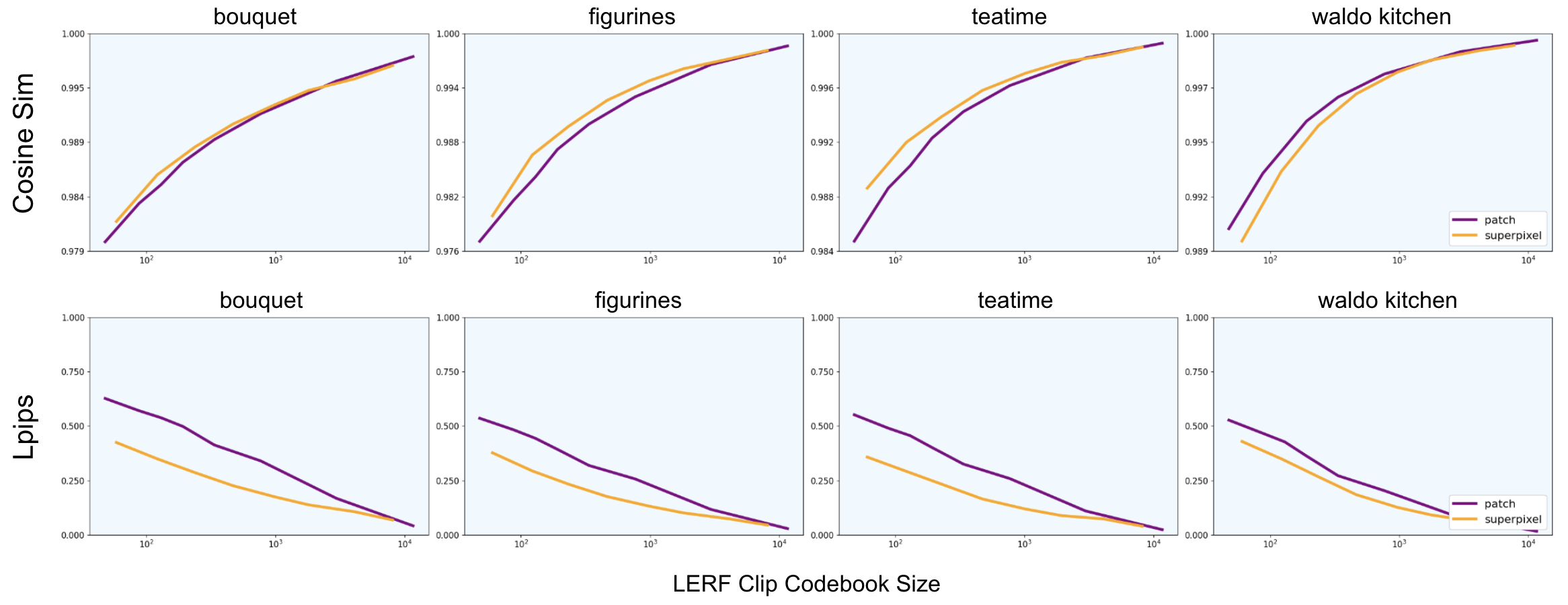}
    \caption{Superpixel-based and patch-based local quantization yield the same cosine distance to the original feature map, but superpixel-based quantization is superior in reconstructing image features. This is attributed to patch-based quantization overfitting noise present in the original feature maps.}
    \label{fig:feature_quant_metrics}
\end{figure*}

\begin{figure*}[t]
    \centering
    \includegraphics[width=0.95\textwidth]{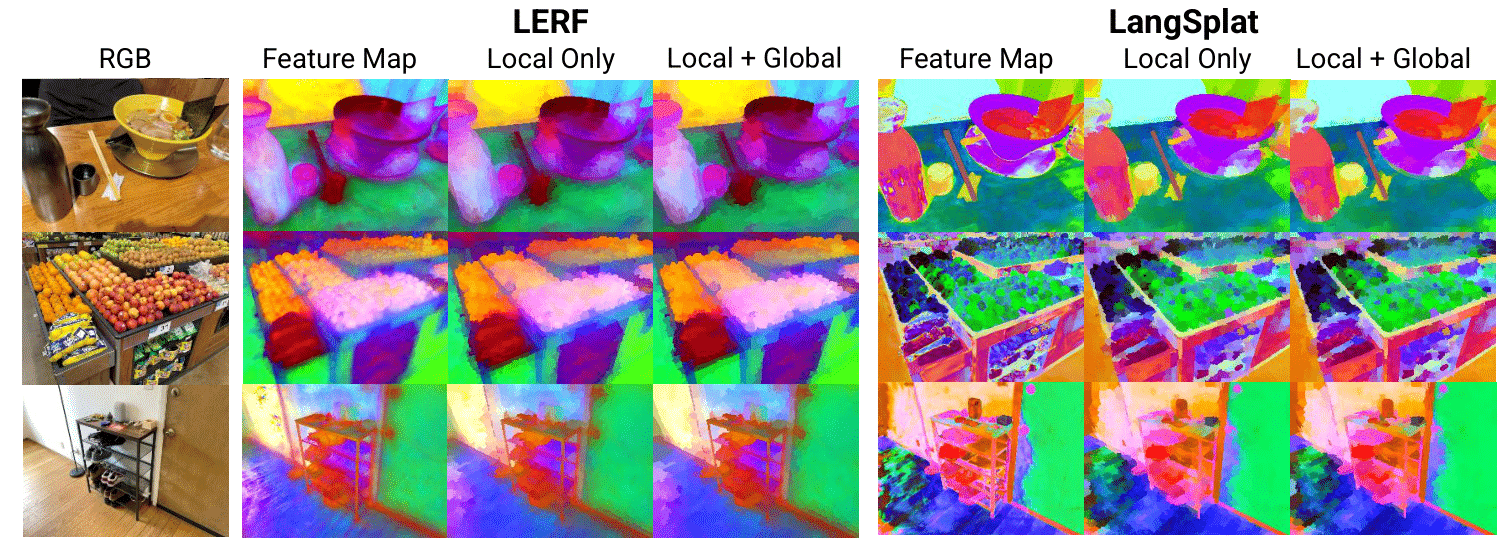}
    \caption{We visualize the original feature maps (left) alongside our proposed local (middle) and local+global (right) alignment methods. Note how our local approach preserves fine-grained spatial information and maintains overall structural consistency with the original features.}
    \label{fig:feature_quant}
\end{figure*}

\begin{figure*}[t]
    \centering
    \includegraphics[width=0.95\textwidth]{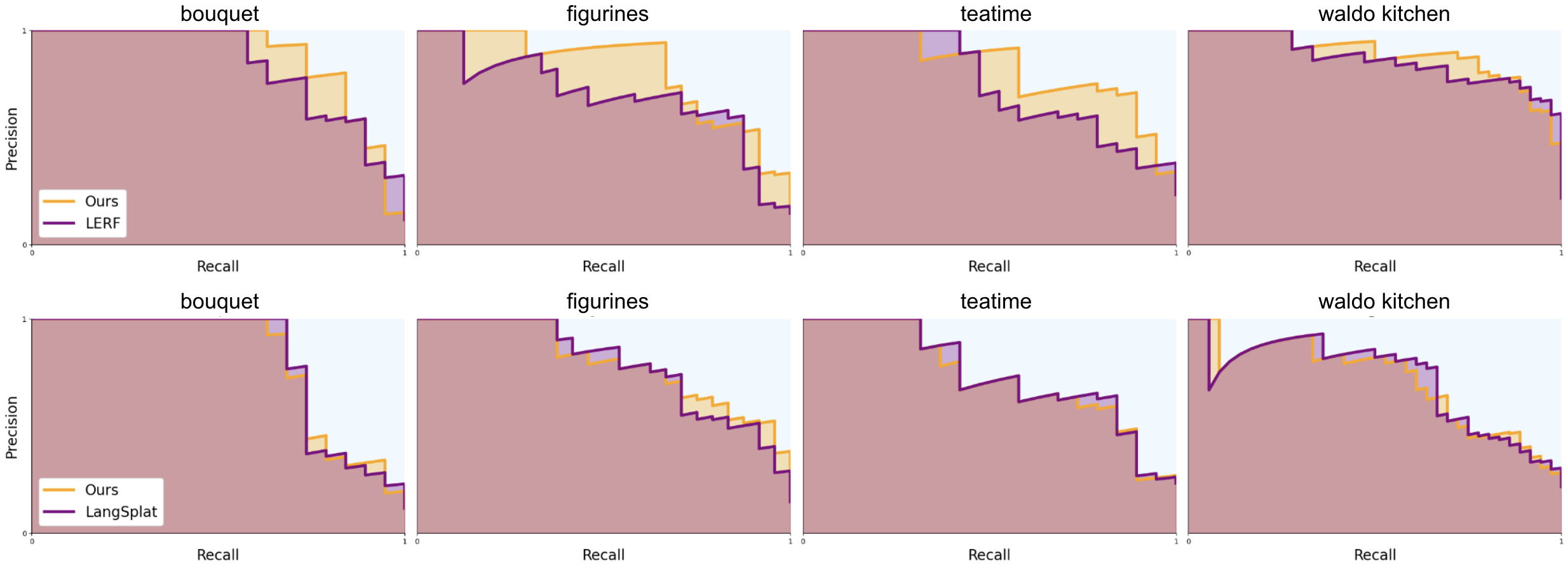}
    \caption{Precision-recall curves comparing our method against LERF and LangSplat for object detection tasks (using LERF's max relevancy measure). Our approach demonstrates consistently higher precision across varying recall thresholds, especially for LERF, which can be attributed to our method's denoising property.}
    \label{fig:object_detection_metrics}
\end{figure*}

\begin{figure*}[t]
    \centering
    \includegraphics[width=\textwidth]{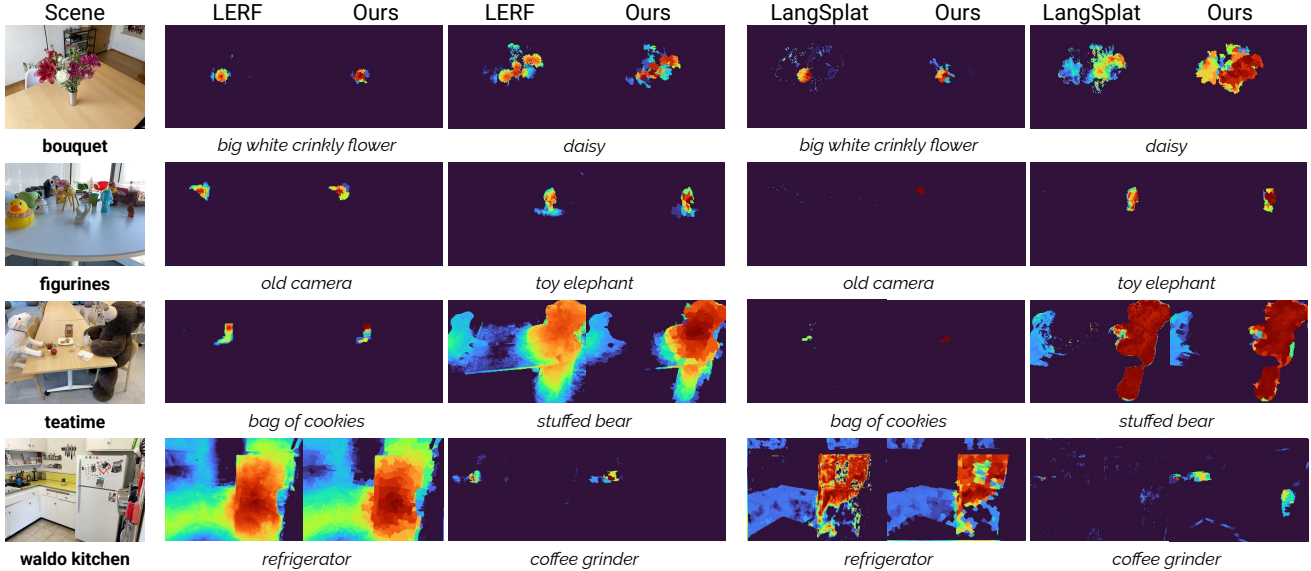}
    \caption{Qualitative comparison of object detection results. Our method often demonstrates more accurate object localization (using LERF's max relevancy measure) and reduced noise compared to LERF (left) and even LangSplat (right). We produce object boundaries that align with that of LERF and LangSplat's despite extensive quantization.}
    \label{fig:object_detection}
\end{figure*}

\subsection{Multistage Vector Quantization}
We describe a multistage vector quantization technique, shown in Figure \ref{fig:method.png}, that preserves feature map fidelity while adhering to the constraint that only one feature map scale can be rendered at once (section 4.1). The first observation is that each feature map can be quantized individually, which we refer to as \emph{local quantization}. Then, during \emph{global quantization}, the quantized components are combined for the final codebook and index map.

\paragraph{Local Quantization} We observe image superpixels, as in \ref{fig:feature_superpixel_alignment}, tend to provide a good partitioning of the feature map for quantization since
\begin{enumerate}[label=(\arabic*)]
    \item Features that are closer in pixels tend to be similar.
    \item The structure of feature maps tends to mirror that of the structure of the image due to multiview training.
\end{enumerate}

We verify and show the implications of these claims in Section 4.6. Superpixels have been used in the past for color quantization, local feature computation, and tracking, but their potential for quantizing feature maps remains unexplored. We compute superpixels using FastSLIC
\begin{align}
S = \text{FastSLIC}(x, N_{\text{superpixels}}, \lambda)
\end{align}
for $N_{\text{superpixels}}$ superpixels and compactness parameter $\lambda$. (1) is influenced via $\lambda$ and both (1) and (2) are influenced by $N_{\text{superpixels}}$. All features within a superpixel are quantized to the same codebook index that stores their spherical mean, the optimal value to minimize cosine similarity distance given a collection of normalized embeddings
\begin{align}
c_{s_k} = \frac{\sum_{f_j \in F_{s_k}} f_j}{||\sum_{f_j \in F_{s_k}} f_j||}
\end{align}
where $F_{s_k} = \{f_j, j \in s_k\}$ denotes the features in superpixel $s_k$. Local quantization creates a local codebook for each scale given a frame and concatenates them into an image codebook $C_{i}, 1 \leq i \leq N$. This significantly reduces the number of features we need to process in the subsequent stages. The runtime complexity of local quantization is $O(HWD)$ per feature map for $NM$ feature maps which results in a codebook of size of at most $M|S|$ per image, and the memory complexity of local quantization is $O(HWD)$.

\paragraph{Global Quantization}
We perform feature pooling by concatenating the local codebooks from all frames and running k-means with the number of clusters 
\begin{align}
K = \alpha|C_{pooled}| \leq \alpha NM|S|
\end{align}
for some constant $\alpha$ and $C_{pooled} = C_1 \oplus C_2 \oplus ... \oplus C_N$ where $\oplus$ denotes concatenation. This global codebook effectively represents the entire scene with a compact set of feature vectors. Global quantization takes 
\begin{align}
O(KD|C_{pooled}|) 
&= O(\alpha D|C_{pooled}|^2)\\
&\leq O(\alpha DN^2M^2|S|^2)
\end{align}
and uses memory according to the k-means implementation.

\subsection{Semantic Lifting Improves Lifting}
We now formulate how semantic lifting can be applied to several areas via different masking operations and lifting functions.

\paragraph{Object-centric Editing}
3D scene editing such as \cite{haque2023instructnerf2nerfediting3dscenes} relies on applying per view InstructPix2Pix, which is mask-free, then lifting the results into 3D. However, InstructPix2Pix often produces extraneous edits for a given prompt. We can replace the $x’ = edit(x)$ step with a composition of the original and edited image to ensure only the described region is affected. Figure \ref{splash} showcases examples.
\begin{align}
x’ = (1 - m_i) \odot x + m_i \odot edit(x)
\end{align}

\paragraph{Efficient OpenEQA}
The best-performing baselines in OpenEQA uniformly strides the sequence and feeds the image subset along with the prompt to a VLM. However, most questions only require a semantically relevant subset of the scenes. For example, asking ``Is the microwave door closed?" only requires the frames with the microwave. We can formulate these localized questions as
\begin{align}
R = \text{VLM}(\text{select}(\{\mathcal{I}\left[|m_i| > P\right] \odot x_i, p_i\}))
\end{align}
where $\mathcal{I}\left[ \cdot \right]$ produces an indicator mask determining whether the masked region is larger than $P$ pixels and the $\text{select}$ function aggregates non-zero masks. \Cref{tab:emeqa_coarse} shows not only can our semantic lifting formulation make agent-based EQA much more efficient, but also more accurate. We provide more details and per-category question splits in the supplemental.

% \paragraph{Localization}
% Often, we don't have the depth of an object in a scan. Monocular depth estimation from a single image can be biased. We can formulate localization from a multiview perspective. Suppose we are given depth maps $d_i$ for each of the $i$ views. Then we can formulate localization via semantic lifting as
% $$p = \text{vote}(\text{select}(\{m_i \odot x_i, p_i\}))$$
% and $\text{vote}$ produces a point in 3D space based on each camera's parameters as well as the depth of each relevant pixel.

\section{Experiments}
\noindent\textbf{Overview.}
We evaluate our VQ-FF approach on a diverse set of 3D scenes, demonstrating superior performance in feature preservation, object detection accuracy, and computational efficiency. Our experiments show that VQ-FF not only maintains feature fidelity but also provides inherent denoising benefits that improve downstream task performance.

% \textbf{Setup}
% We evaluate our method on the 14 scenes from the LERF Dataset, which includes a diverse selection of complicated scenes ranging from object clutters and indoor rooms to grocery stores and outdoor environments. To showcase the generality of our method, we benchmark using LERF (pixel-aligned features, 30 scales, NERF) and LangSplat (part-aligned features, 3 scales based on Segment Anything, Gaussian Splat) as the grounding scene representations. We train the LERF and LangSplat fields using the default settings given in \cite{kerr2023lerflanguageembeddedradiance} and \cite{qin2024langsplat3dlanguagegaussian}.

\noindent\textbf{Experimental Setup.}
We conduct experiments on 14 scenes from the LERF Dataset, encompassing diverse environments from cluttered indoor rooms to outdoor scenes. To demonstrate the generality of our approach, we evaluate using two different scene representations:
\begin{itemize}
    \item LERF: Uses pixel-aligned features with 30 scales in a NeRF framework
    \item LangSplat: Employs part-aligned features with 3 scales based on Segment Anything in a Gaussian Splatting framework
\end{itemize}
Both representations are trained using default parameters from their respective papers \cite{kerr2023lerflanguageembeddedradiance, qin2024langsplat3dlanguagegaussian}.

\noindent\textbf{Feature Map Alignment Analysis.}
We first analyze our multistage quantization approach

\noindent\emph{Local Quantization.} 
We evaluate quantization error across feature maps at all scales, sampling at stride $s = \lfloor \frac{N}{n} \rfloor$ with $n = 20$ images per sequence. Our baseline comparison uses the spherical mean of image patches, which theoretically minimizes cosine distance within each patch. We examine two metrics:
\begin{itemize}
    \item Cosine distance ($\uparrow$): Comparing reconstruction error across codebook sizes
    \item LPIPS ($\downarrow$): Evaluating structural alignment via PCA visualization
\end{itemize}
We test patch sizes $p \in [2, 4, 8, 12, 16, 20, 24, 32]$, and $N_{superpixels}$ $\in [8192, 4096, 2048, 1024, 512, 256, 128, 64]$.

Our findings, detailed in Figure \ref{fig:feature_quant_metrics}, show that while both quantization methods maintain similar cosine distances, superpixel quantization significantly improves LPIPS scores. This improvement extends to both Dino features and LangSplat, where we observe enhanced structural alignment and slight improvements in cosine distance (see supplemental).

The divergence between cosine distance and LPIPS metrics stems from the fundamental differences in quantization approaches. Patch-based methods, being location-dependent rather than content-aware, may better capture noise while sacrificing edge alignment. In contrast, our superpixel-based approach provides natural denoising by respecting image structure, leading to improved feature map quality. This is evident in Figure \ref{fig:feature_quant_local_lerf}.

\noindent\emph{Global Quantization.}
We assess feature map consistency across all scales and scenes using stride $s$. Given the high correlation between Clip embeddings, we use the spherical mean feature map distance as a scale reference. Our quantized feature maps demonstrate strong alignment with ground truth, as shown by Figure \ref{fig:feature_quant} and Table \ref{tab:lerf_comparison}.

\begin{table}[h]
    \centering
    \adjustbox{max width=\linewidth}{
    \begin{tabular}{lrr}
    \toprule
        Method & Reference & Ours \\
    \midrule
        LERF (Clip) & 0.941 & \textbf{0.995} \\ 
        LERF (Dino) & 0.896 & \textbf{0.977} \\ 
        LangSplat (Clip) & 0.556 & \textbf{0.902} \\ 
    \bottomrule
    \end{tabular}
    } % end adjustbox
    \caption{\textbf{Feature map similarity comparison between our VQ-FF quantization and reference methods}. Higher values indicate better preservation of feature information.}
    \label{tab:lerf_comparison}
\end{table}

\noindent \textbf{Object Detection.}
We now benchmark querying on the LERF object section dataset, a subset of LERF with object localization annotations for each scene. We plot the precision and recall curves for LERF and LangSplat, where the baselines are computed by rendering and querying every single feature map in the sequence. We follow LERF and compute metrics based on how often the max relevancy in a given image falls within a labeled bounding box. As noted above and as seen in \Cref{fig:object_detection_metrics}, due to the denoising property of superpixel quantization, our $O(1)$ codebook search and index method slightly exceeds LERF in precision given a fixed recall, which can be attributed to fewer false positives due to denoising property and is on par with LangSplat. Note LangSplat performs worse than LERF due to less granularity, which is why the performance boost due to less noise is not observed. \Cref{fig:object_detection} compares relevancy maps produced by LERF and LangSplat with VQ-FF.

\noindent \textbf{Performance.}
Under our chosen parameters, LERF scenes (30 scales, both Dino and Clip) take on average 6m45s to quantize, while LangSplat (3 scales based on SAM, Clip) take on average 34s to quantize (\Cref{tab:render_time}).

Per-image both LERF and LangSplat are very efficient, with full sequences of under 400 images taking a few GB of memory. \Cref{tab:memory_usage} shows per-image memory requirements.

LERF Clip codebook obtains bits/dim of 17.62 and Dino codebook bits/dim of 17.95. LangSplat obtains bits/dim of 13.21, since due to SAM mask-based training, LangSplat features are lower granularity and less noisy, making them easier to compress.

\begin{table}[h]
    \centering
    \adjustbox{max width=\linewidth}{
    \begin{tabular}{lr}
    \toprule
        Renderer & Seconds \\
    \midrule
        LERF & 6.748 \\ 
        LangSplat & 0.567 \\ 
    \bottomrule
    \end{tabular}
    } % end adjustbox
    \caption{\textbf{Per-frame quantization time comparison.} Both LERF and Langsplat are processed efficiently, with runtimes scaling according to the number of scales.}
    \label{tab:render_time}
\end{table}

\begin{table}[h]
    \centering
    \adjustbox{max width=\linewidth}{
    \begin{tabular}{lrrr}
    \toprule
        Renderer & Codebook & Codebook Indices & Combined \\
    \midrule
        LERF (Clip) & 2.891 & 5.251 & 8.142 \\
        LERF (Dino) & 0.072 & 0.175 & 0.247 \\
        LangSplat (Clip) & 0.288 & 0.514 & 0.802 \\
    \bottomrule
    \end{tabular}
    } % end adjustbox
    \caption{\textbf{Memory footprint breakdown} (MB per frame) showing storage requirements for different components. Both LERF variants and LangSplat achieve compact representation.}
    \label{tab:memory_usage}
\end{table}

\begin{table}[h]
    \centering
    \adjustbox{max width=\linewidth}{
    \begin{tabular}{lrrr}
    \toprule
        Method & LLM-Match & Frame Reduction \\
    \midrule
        GPT4-V & $51.3 \pm 1.8$ & ----- \\
        GPT4-V (VQ-FF) & \textbf{$52.9 \pm 1.9$} & \textbf{1.8x} \\
    \bottomrule
    \end{tabular}
    } % end adjustbox
    \caption{\textbf{EM-EQA performance comparison} between GPT-4V (Baseline) and GPT-4V (VQ-FF) in terms of LLM-Match accuracy and the number of frames used for querying, on the HM3D dataset. The baseline uses 50 uniformly strided frames from the image sequence.}
    \label{tab:emeqa_coarse}
\end{table}
\section{Conclusion}
In this work, we introduced semantic lifting, a novel framework that generalizes traditional lifting tasks by incorporating per-image masks to identify relevant regions. We demonstrated how our Vector-Quantized Feature Field (VQ-FF) enables efficient and accurate semantic lifting through a two-stage quantization process that preserves feature fidelity while drastically reducing memory requirements. Our approach not only maintains or improves query accuracy compared to existing methods but also enables new applications in object-centric editing and efficient embodied question answering. Through extensive experiments, we showed that our method achieves superior precision in object detection tasks while requiring significantly less memory than previous approaches, making it a practical solution for large-scale 3D scene understanding and manipulation tasks.
{
    \small
    \bibliographystyle{ieeenat_fullname}
    \bibliography{main}
}

\clearpage

\pagenumbering{arabic}
\renewcommand*{\thepage}{A\arabic{page}}

\setcounter{figure}{0}
\renewcommand\thefigure{S\arabic{figure}}   

\setcounter{table}{0}
\renewcommand\thetable{S\arabic{table}}   

\onecolumn
\begin{center}
    {\Large \textbf{Supplementary Material}}\\[10mm]
\end{center}

\appendix

\renewcommand{\thesection}{\Alph{section}}

\section{Additional Results for Localized Editing}
We showcase more examples of semantic lifting for object-centric editing. We apply InstructPix2Pix for editing. The results were lifted using Gaussian Splatting. The top row of each figure indicates naively lifting InstructPix2Pix edits while the bottom row semantic lifting.
\\
\begin{figure*}[!h]
    \centering
    \includegraphics[width=0.75\textwidth]{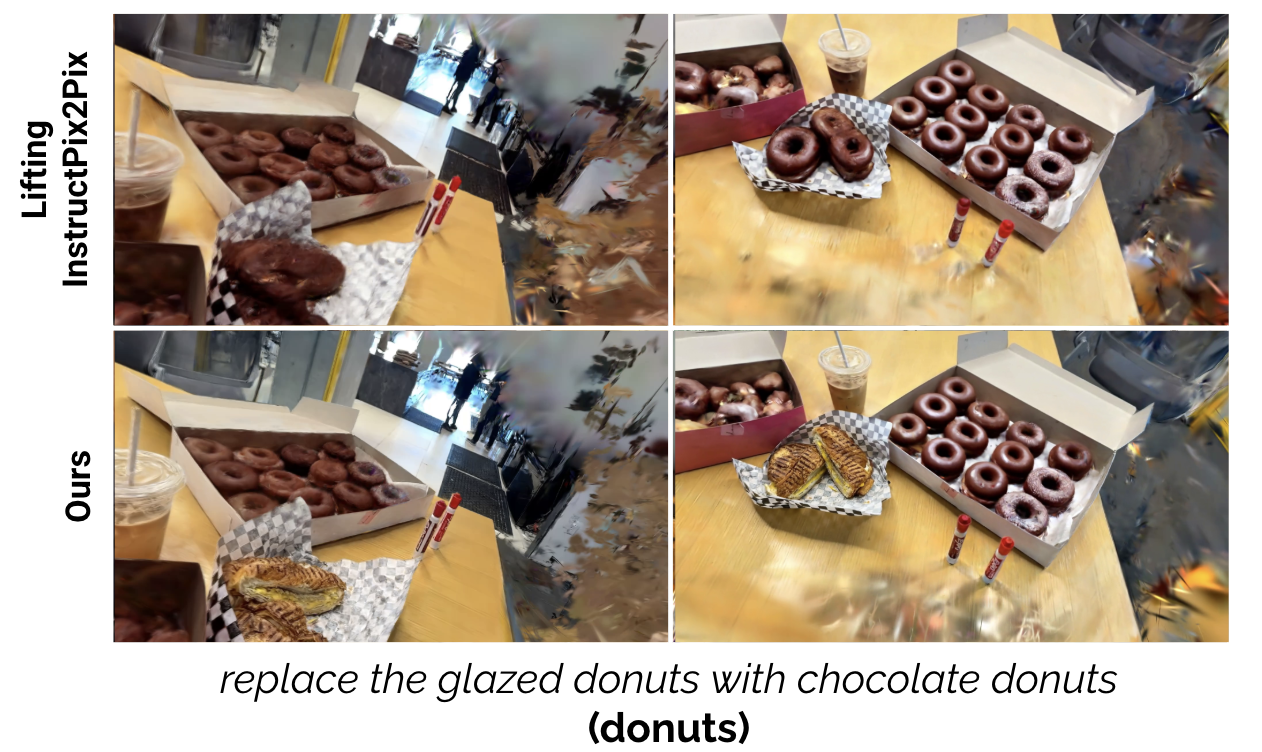}
    \caption{Observe the sandwich is corrupted into a donut without semantic lifting. Multiple iterations in an InstructNeRF2NeRF paradigm will not correct this type of error.}
\end{figure*}

\begin{figure*}[!h]
    \centering
    \includegraphics[width=0.75\textwidth]{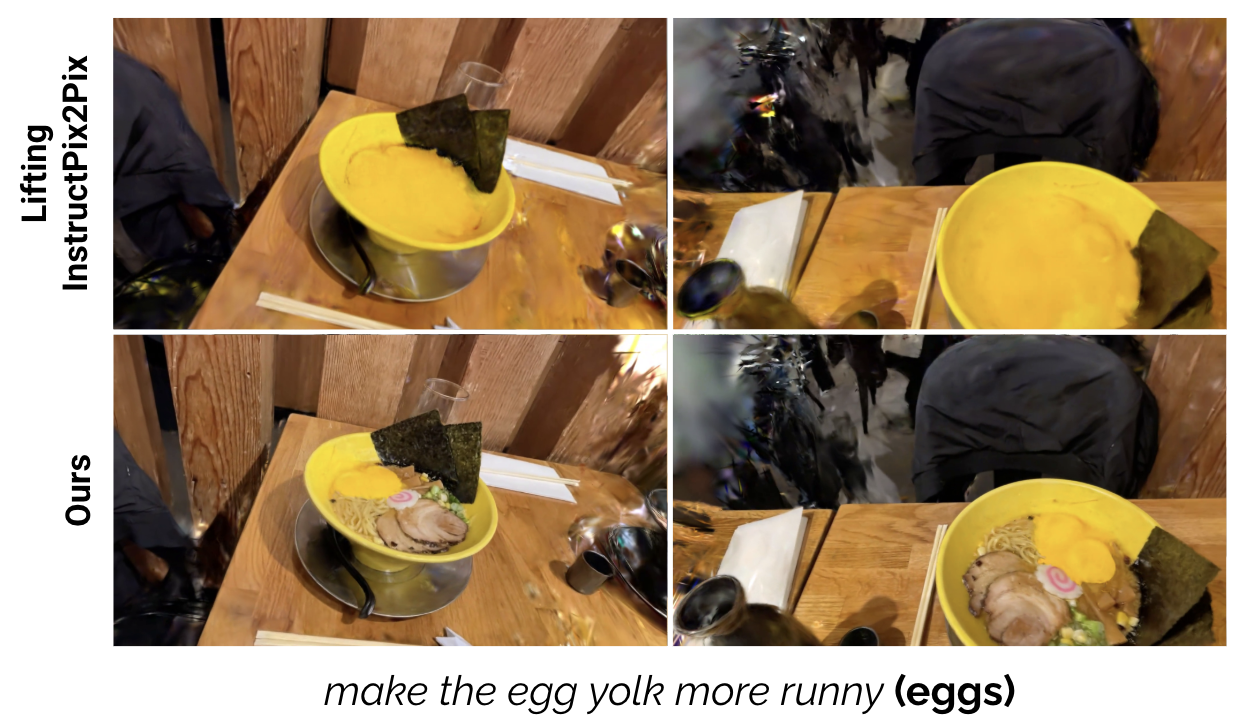}
    \caption{Similar to the donuts scene, the entire ramen bowl is edited as opposed to only the eggs.}
\end{figure*}

For certain types of edits that are hard to describe and out of distribution for instruction models such as \textit{Modify the depiction of the fruits so they reflect the impressionistic style characteristic of Monet.}, we can formulate using semantic lifting e.g. change the images to be in the style Monet (which InstructPix2Pix is great at) and then apply masking so only the fruits appear edited.

\begin{figure*}[!h]
    \centering
    \includegraphics[width=0.75\textwidth]{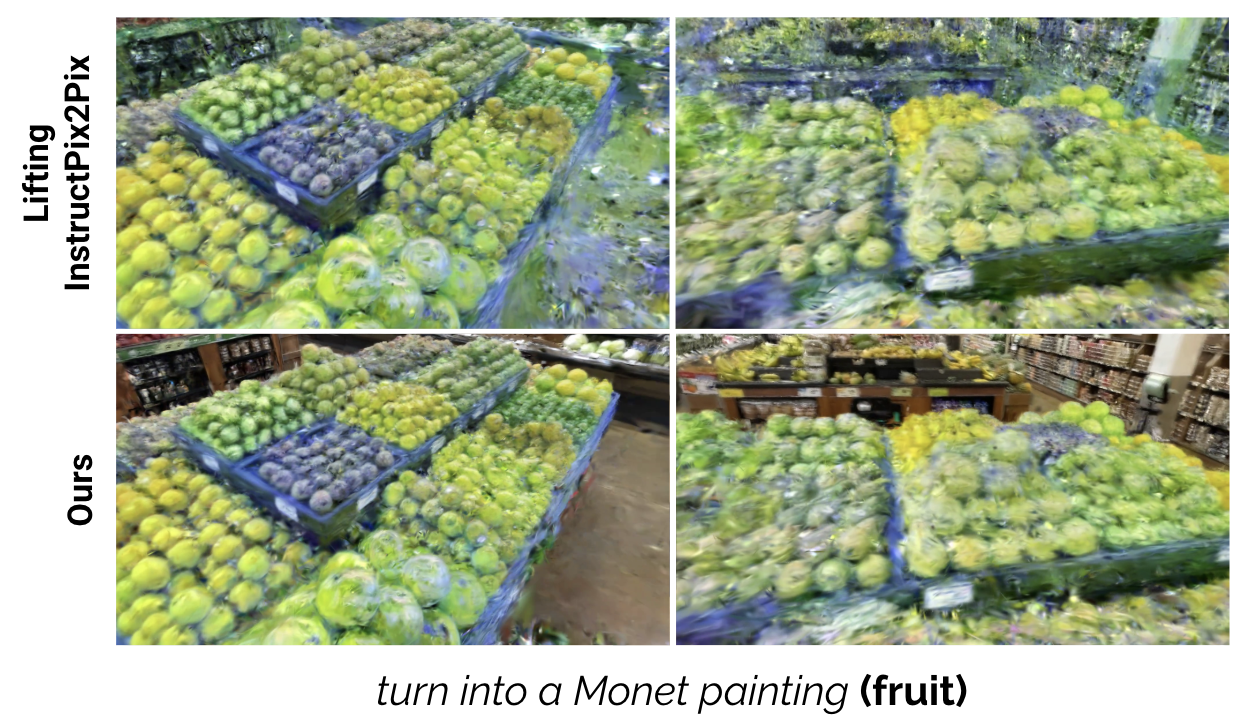}
    \caption{Certain difficult to describe editing scenarios can be easily formulated with semantic lifting.}
\end{figure*}

\section{OpenEQA Benchmark}
We evaluate our approach on the OpenEQA benchmark, an embodied question answering (EQA) dataset, which evaluates large-models on their ability to answer questions about scenes. Specifically, we compare against the GPT-4V benchmark on the HM3D dataset consisting of 537 questions, which takes in a subset of $50$ scene images along with the question as input. In the original OpenEQA baseline, the frames are randomly selected without considering the relevance of question-specific content. In contrast, our approach employs an intelligent frame-selection strategy by Clip similarity matching. First, we extract a Clip token from the provided question by querying a large language model (GPT-4o) with the following high-level context, and a few examples. \begin{quote}
You are a CLIP token generator for vision-language retrieval tasks. Given a natural language question about a scene, extract the most relevant keywords and reformat them into a concise, CLIP-friendly query. Follow these rules:
\begin{itemize}
\item Keep only important objects, locations, and attributes.
\item Remove filler words like `what', `is', `the', `in', `on', `above', etc.
\item Do not include spatial associations or relationships (e.g., `chest sofas' instead of `chest color surrounded by sofas?').
\item DO NOT return full sentences; only a short, descriptive phrase.
\end{itemize}
\end{quote}

\begin{quote}
For example:\\
Q: What is the white object on the wall above the TV? \newline
A: white object TV \newline
Q: What color is the staircase railing? \newline
A: staircase railing \newline
Q: What is in between the two picture frames on the blue wall in the living room? \newline
A: picture frames blue wall living room 
\end{quote}

Using these CLIP tokens, we compute the relevancy masks between the token embeddings and frame embeddings using a CLIP encoder. We use a threshold of $10\%$ to filter irrelevant frames, ensuring the relevance of the chosen images to the query. To ensure diversity, frames are grouped into two sets according to the area of their relevancy masks, consisting of the top $50\%$ and bottom $50\%$ frames. From each set, we select up to $25$ frames using a uniform sampling strategy, ensuring a maximum of $50$ selected frames. The selected frames are then inputted to GPT-4V, along with the original question, to generate answers. Although in practice, we observe that our method requires $1.8\times$ fewer frames than the baseline of $50$ frames, as irrelevant frames are effectively filtered out.

In Table~\ref{tab:emeqa_fine}, we show the breakdown of LLM-Match scores between the model-generated answer and the human-annotated answer across the 7 question categories, specified in the OpenEQA benchmark. 

\begin{table}[h]
    \centering
    \begin{adjustbox}{max width=\linewidth}
    \begin{tabular}{lccccccc}
    \toprule
        \multirow{2}{*}{Method} & \makecell{Object \\ Recognition} & \makecell{Object \\ Localization} & \makecell{Attribute \\ Recognition} 
        & \makecell{Spatial \\ Understanding} & \makecell{Object State \\ Recognition} & \makecell{Functional \\ Reasoning} & \makecell{World \\ Knowledge} \\
    \midrule
        % GPT-4V & 2.891 & 5.251 & 33 & 33 & 33 & 33 & 33 \\
        GPT-4V (VQ-FF) & 50.1 & 51.3 & 54.5 & 41.7 & 56.2 & 64.6 & 52.1 \\
    \bottomrule
    \end{tabular}
    \end{adjustbox}
    \caption{EM-EQA performance comparison between GPT-4V (Baseline) and GPT-4V (VQ-FF) in terms of LLM-Match accuracy and the number of frames used for querying, on the HM3D dataset across different question categories. The baseline uses 50 uniformly strided frames from the image sequence.}
    \label{tab:emeqa_fine}
\end{table}

\section{Local Quantization Results for LERF Dino and Langsplat Clip}
We show results for LERF Dino and LangSplat Clip local quantization. Similar to LERF Clip local quantization, we see similar performance in cosine similarity but better performance for superpixel-based quantization in terms of Lpips. This is because patch-based quantization, due to fitting according to position locality, better fits noise. We see that for a less noisy grounding scene representation, such as LangSplat, superpixel-based quantization also slightly exceeds patch-based quantization in terms of cosine distance.
\\
\\
\begin{figure*}[!h]
    \centering
    \includegraphics[width=\textwidth]{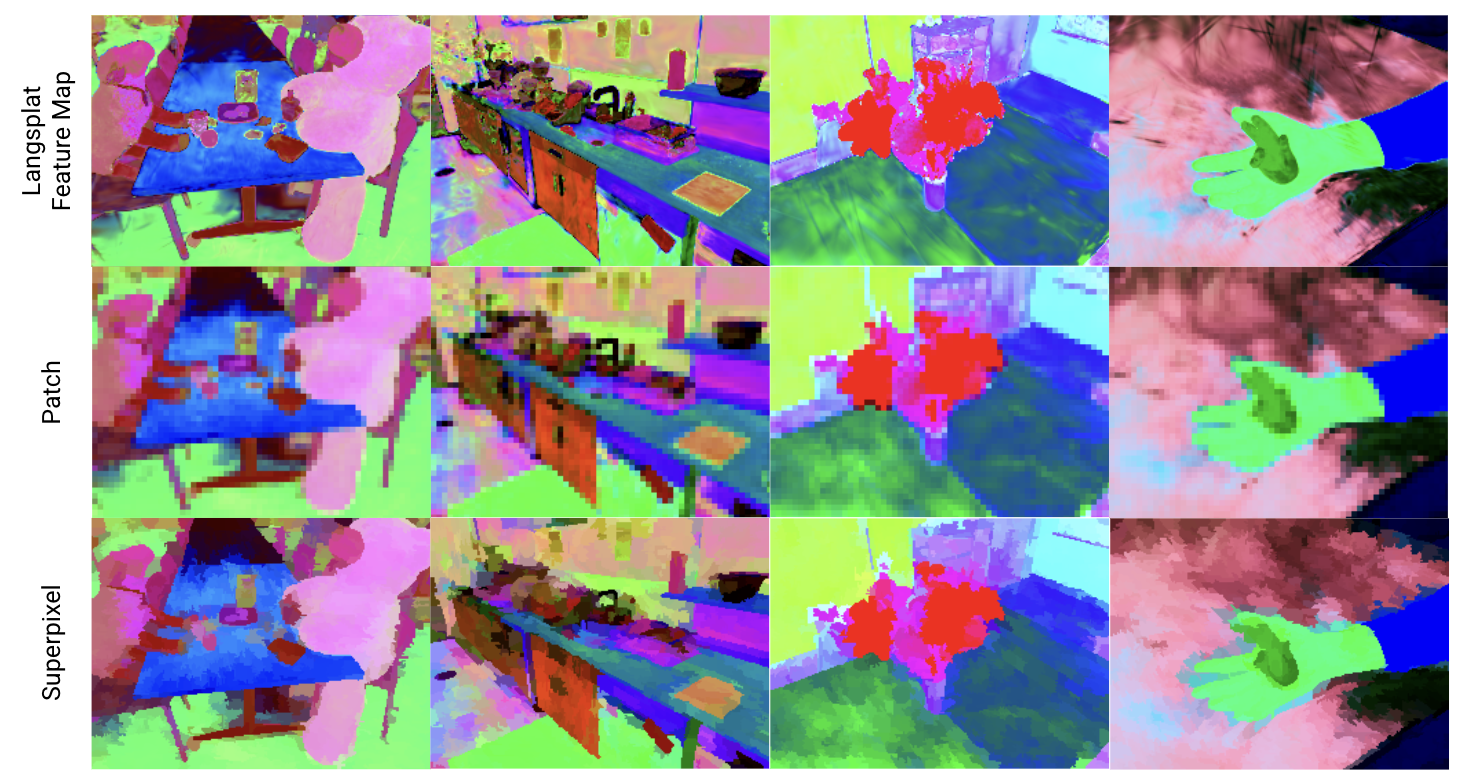}
    \caption{For Langsplat Clip features, superpixel-based local quantization has better cosine similarity and lower Lpips.}
\end{figure*}

\begin{figure*}[!h]
    \centering
    \includegraphics[width=\textwidth]{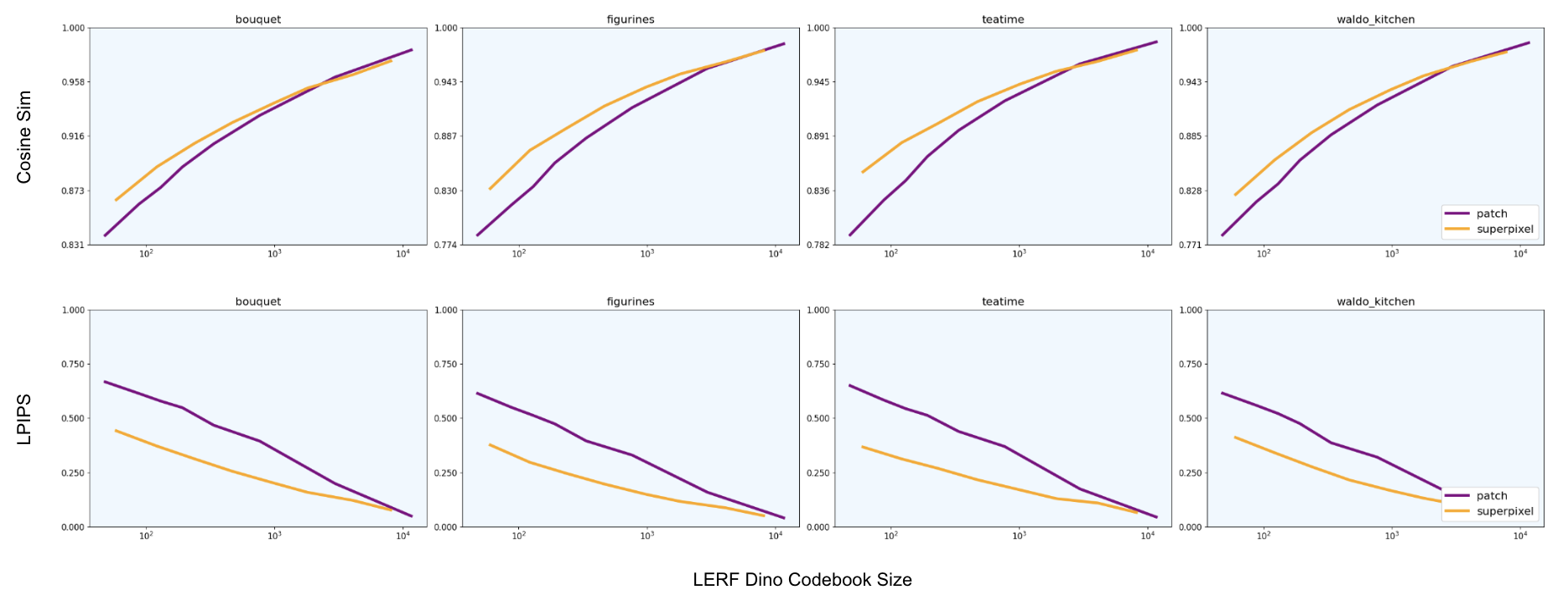}
    \caption{Like for LERF Clip features, superpixel-based local quantization has similar cosine distance but greatly reduced Lpips.}
\end{figure*}

\begin{figure*}[!h]
    \centering
    \includegraphics[width=\textwidth]{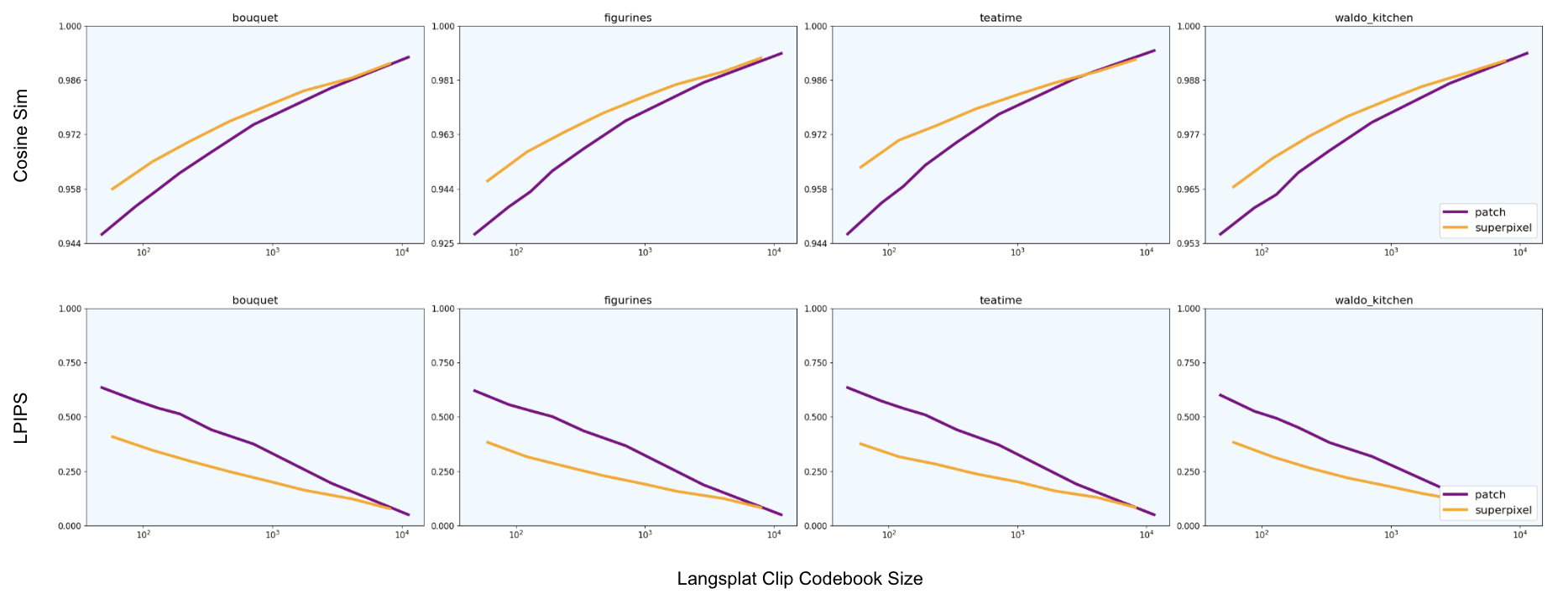}
    \caption{We visualize LangSplat Clip features for superpixel and patch-based local quantization. Observe superpixel-based quantization is superior at reconstructing boundaries and preserving image structure.}
\end{figure*}

\clearpage
\section{Additional Local Quantization Results}
We now present additional results for local quantization for the remaining scenes in the LERF dataset and observe similar trends.

\begin{figure*}[!h]
    \centering
    \includegraphics[width=\textwidth]{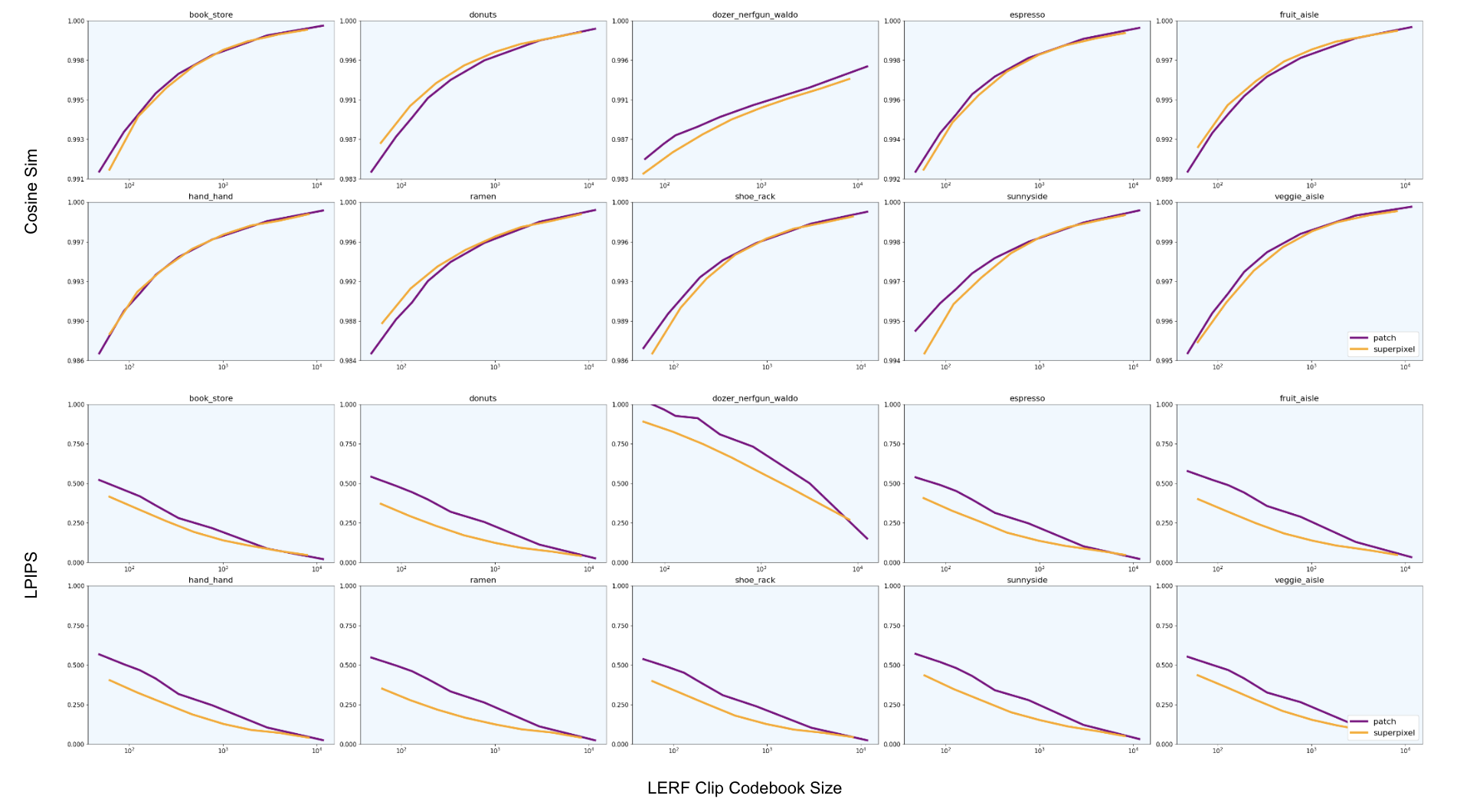}
    \caption{Additional LERF Clip metrics.}
\end{figure*}

\begin{figure*}[!h]
    \centering
    \includegraphics[width=\textwidth]{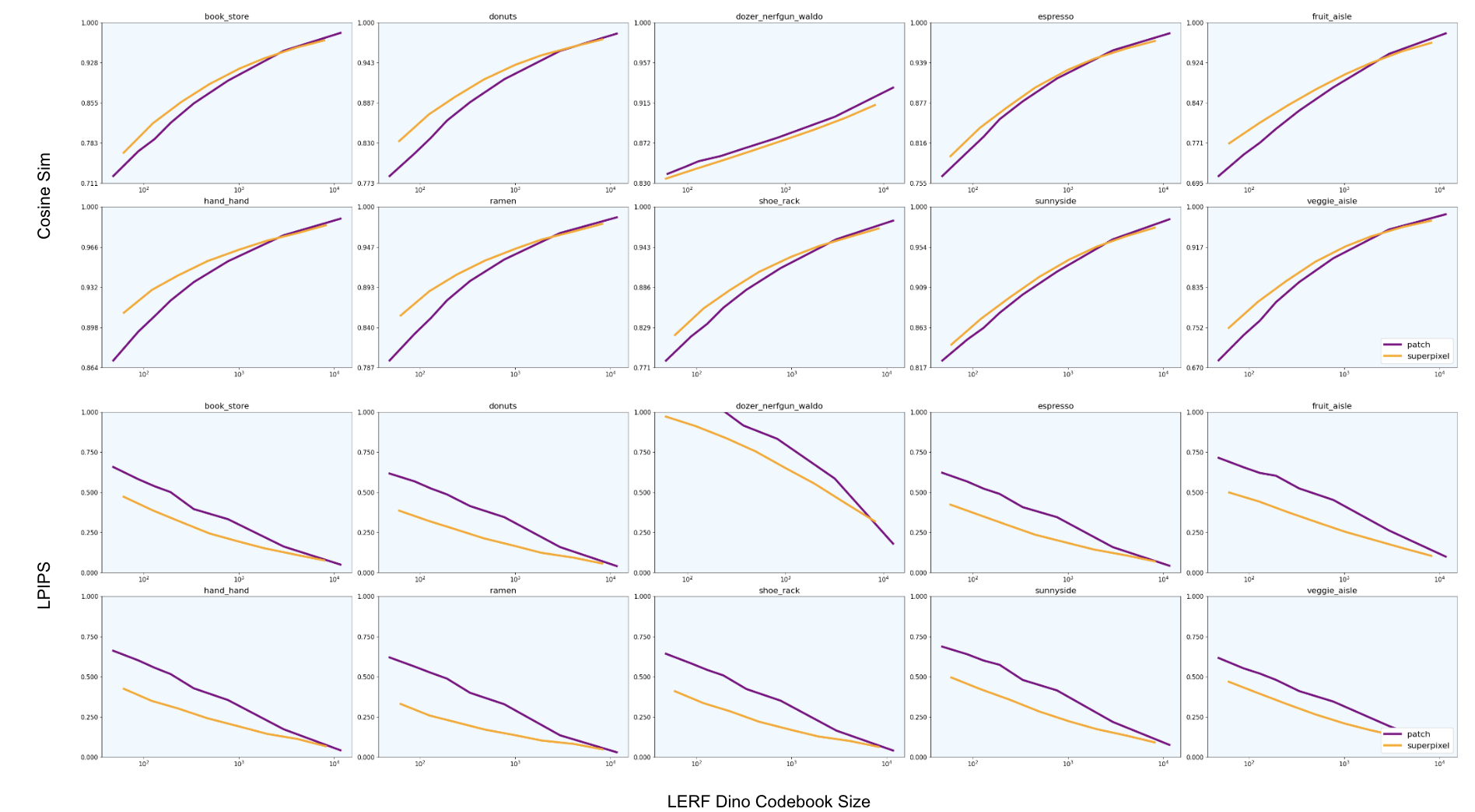}
    \caption{Additional LERF Dino metrics.}
\end{figure*}

\begin{figure*}[!h]
    \centering
    \includegraphics[width=\textwidth]{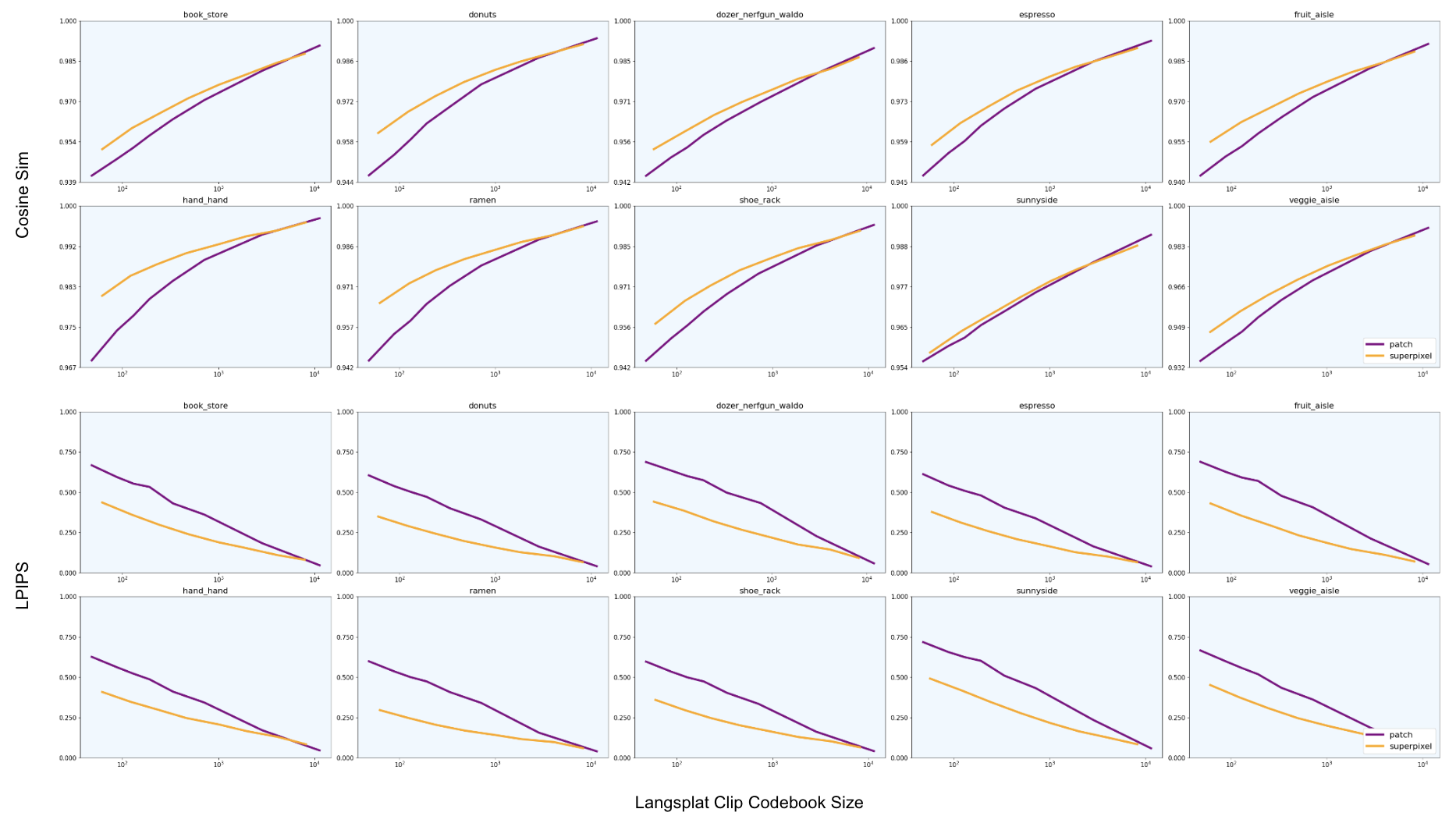}
    \caption{Additional Langsplat Clip metrics.}
\end{figure*}

\clearpage
\section{Additional Quantization Visualizations}
We show additional local and global quantizations of LERF and LangSplat feature maps. 
\begin{figure*}[!h]
    \centering
    \includegraphics[width=\textwidth]{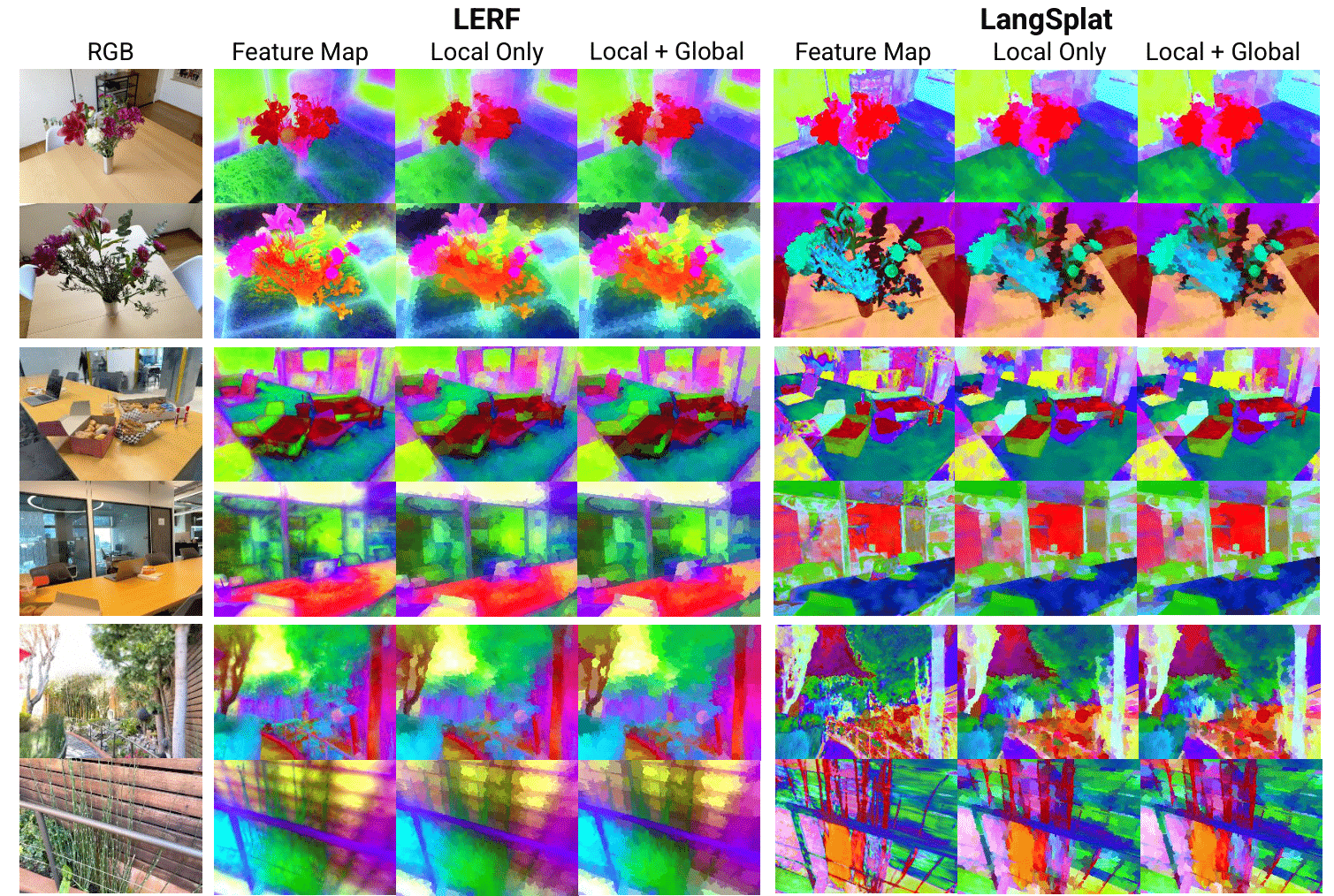}
\end{figure*}

\end{document}